\documentclass[runningheads]{llncs}
\usepackage[T1]{fontenc}
\usepackage{graphicx,subfigure}
\usepackage{amsmath,amssymb} % define this before the line numbering.

\usepackage{color}
\usepackage{marvosym}
\usepackage{hyperref}
\usepackage{bm}
\usepackage{multirow}
\usepackage{enumitem}   
\usepackage[ruled]{algorithm2e}
\usepackage{algorithmic}

\newcommand{\ie}{\textit{i.e.}}
\newcommand{\eg}{\textit{e.g.}}
\newcommand{\bb}{\mathbf{b}}
\newcommand{\pp}{\mathbf{p}}
\newcommand{\lp}{\mathbf{LP}}
\newcommand{\xx}{\mathbf{x}}
\newcommand{\XX}{\mathbf{X}}
\newcommand{\EE}{\mathbf{E}}

\newcommand{\maxx}{\operatorname{max}}
\newcommand{\diag}{\operatorname{diag}}

\newcommand{\func}[1]{\textcolor{blue}{\mathtt{#1}}}

\newcommand{\variable}[1]{\textcolor{black}{\mathtt{#1}}}
\newcommand{\param}[1]{\textcolor{red}{\mathtt{#1}}}

\definecolor{blue}{RGB}{60,132,196}
\definecolor{red}{RGB}{207,78,56}
\definecolor{gray}{RGB}{146,146,161}

\begin{document}
\title{Weighted Contrastive Hashing}
%
%\titlerunning{Weighted Contrastive Hashing}
\author{Jiaguo Yu \and
Huming Qiu \and
Dubing Chen \and
Haofeng Zhang \textsuperscript{(\Letter)}}
\authorrunning{J. Yu et al.}
\institute{School of Computer Science and Engineering, Nanjing University of Science and Technology, Nanjing 210094, China. \\
\email{\{yujiaguo, 120106222682, db.chen, zhanghf\}@njust.edu.cn} }

\maketitle              % typeset the header of the contribution
\begin{abstract}
The development of unsupervised hashing is advanced by the recent popular contrastive learning paradigm. However, previous contrastive learning-based works have been hampered by (1) insufficient data similarity mining based on global-only image representations, and (2) the hash code semantic loss caused by the data augmentation. In this paper, we propose a novel method, namely Weighted Contrative Hashing (WCH), to take a step towards solving these two problems. We introduce a novel mutual attention module to alleviate the problem of information asymmetry in network features caused by the missing image structure during contrative augmentation. Furthermore, we explore the fine-grained semantic relations between images, \ie, we divide the images into multiple patches and calculate similarities between patches. The aggregated weighted similarities, which reflect the deep image relations, are distilled to facilitate the hash codes learning with a distillation loss, so as to obtain better retrieval performance. Extensive experiments show that the proposed WCH significantly outperforms existing unsupervised hashing methods on three benchmark datasets.
Code is available at: \href{http://github.com/RosieYuu/WCH}{\textcolor[RGB]{0,0,255}{http://github.com/RosieYuu/WCH}}.
\end{abstract}
\section{Introduction}
With the advancement of deep neural networks, deep hash has become one of the most studied approaches for Approximate Nearest Neighbors (ANN) in large-scale image retrieval. Earlier studies rely heavily on artificial annotations, which makes it difficult to apply in real-world scenarios due to the high labor costs. As a result, unsupervised deep hashing~\cite{cibhash,date,cimon,nsh} has gradually become the major research direction in this field, with the recent boom in unsupervised learning \cite{simclr,moco,mocov2,siamese,swav,byol}. The key difficulty with unsupervised hash is that the ad-hoc encoding process does not extract the key information for hashing, precisely because of the lack of supervised information. Hence, numerous methods have been proposed to learn better discrete representations for hashing in unsupervised setting.

A large family of recent unsupervised hash learning tasks is based on contrastive learning \cite{simclr,moco,mocov2}. These methods build upon instance discrimination, which constructs similar and dissimilar instances and learns the discrete representations by prompting the model to pull in the similar instances and push away the dissimilar instances. With simply the most fundamental concepts for contrastive learning, existing methods\cite{cibhash,cimon,nsh} based on contrastive learning have achieved significant success.

\begin{figure}[!t]
	\begin{center}
		\includegraphics[width=0.65\linewidth]{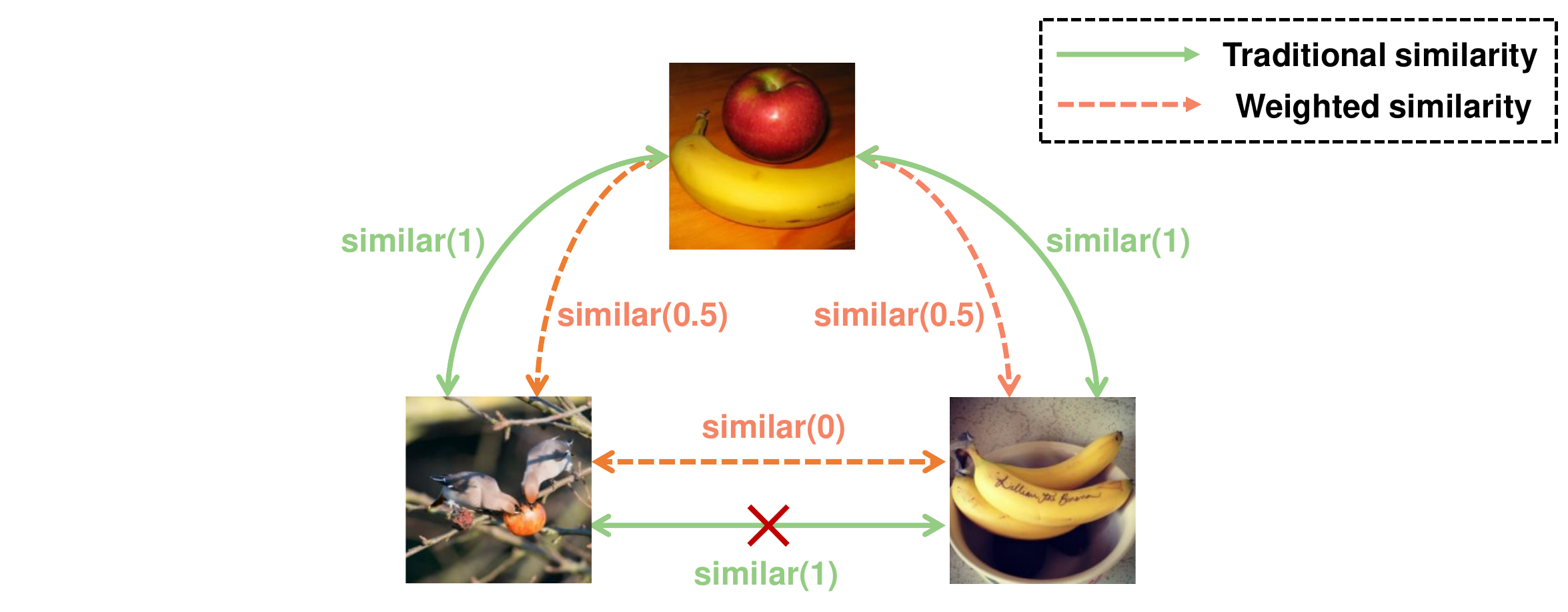}
	\end{center}
	\caption{An example of the conflict of the traditional similarity calculation approach. A typical unsupervised method will treat the top image with both the bottom-left and bottom-right images as similar pairs, because they have a common label. In this case, the two images in the bottom left and bottom right corners should be considered as a similar sample pair. However, the fact is that these two images do not have a common label and they should be considered as a dissimilar pair.}
	\label{fig_1}
\end{figure}

Despite their success, most of the current methods mainly focus on adjusting the contrastive loss to fit the hash learning criterion~\cite{cibhash,cimon}. However, directly combining contrastive loss and unsupervised hashing tasks like this leads to two problems. On the one hand, an instance discrimination-based approach leads to the fact that even if the samples are very similar, they still need to be forced apart, \ie, the sample similarity obtained in this way is unreliable. On the other hand, calculating the similarity with the feature vector or hash code of a whole image may lead to the following problem: The top image in Fig. \ref{fig_1} is associated with the labels of apple and banana; the bottom left image is associated with the labels of apple and bird; and the bottom right image is associated with the labels of banana and bowl. In the traditional method, the similarity of both the bottom left image and the bottom right image according to the top image is considered full similar. Therefore, we can say that the bottom left and bottom right images are also very similar. However, the labels of the bottom left and bottom right images do not overlap, \ie, they are actually dissimilar. Based on this, we raise a question: \textit{how to define or even use the similarity between samples to learn high-quality hash codes?}

Curiously, most existing approaches do not focus on this problem. To the best of our knowledge, NSH~\cite{nsh} uses Neural Sorting Operators to obtain the permutation of a vector of similarity scores, and it employs the sorted similarity results to pick the top m positive samples of the anchor, \ie, it improves the comparison by increasing the number of positive samples in the learning framework. However, in the experiment, the optimal number of positive samples is fixed at 3, and all of them are considered fully similar. 
There are two drawbacks. First, the anchor image and the augmented similar image, especially processed with random crop, are not always similar, \eg, the cropped image only contains the background, which will prompt the network to learn the background rather than the object representation in the image. Second, for multi-labeled images, the positions and sizes of objects vary greatly, and it is difficult to learn a single depth representation that fits all objects. The quality of the hash codes obtained in this way is not high, which will have an impact on the final retrieval results. This can also explain why NSH~\cite{nsh} boosts highly on single-label datasets, but the boost of MAP on multi-label datasets is not very obvious.

In order to solve the above problem, we propose a novel method called Weighted Contrastive Hashing (WCH) to re-weight the similarity of the anchor image and the others. Concretely, we develop a novel metric rule that is more reasonable and efficient for measuring similar samples, and finally apply this rule to the learning of hash codes for better retrieval performance. We divide each image into a number of patches, and exploit the Vision Transformer (ViT)~\cite{vit} as the encoder to adapt the patches as the input to the model. To obtain the similar samples of an anchor, we use the aggregated vector of similarity between the patches of different samples as weights. Unlike NSH~\cite{nsh}, we do not selectively pick the most relevant samples as the positive samples for contrastive learning, but assign trainable weights to all candidate samples, which represent the degree of similarity between samples. That is, we can consider an image pair as less or more similar, rather than stating them as fully similar or dissimilar in absolute terms. Notably, we demonstrate in the experimental section that our method works better than NSH~\cite{nsh}. In addition, to solve the problem of insufficient similarity between augmented images and anchor images, we propose a Mutual Attention (MA) module to reset the weights of each patch of them by calculating their similarity, which can guarantee sufficient similarity of them to make them the most similar pair, so as to facilitate the hash code learning towards the correct direction. In a nutshell, our main contributions are summarized as follows:
\begin{itemize}
\item To the best of our knowledge, this is the first time that weighted contrastive learning has been introduced to image retrieval tasks. It alleviates the problem that certain anchor images and negative samples, which are similar enough for the hashing task, are treated as dissimilar pairs.

\item We propose a Mutual Attention module to achieve information complementation between the augmented anchor image and the positive sample, avoiding the lack of key information for hashing.

\item The excellent performance of our WCH model is extensively demonstrated by comparing it to 18 state-of-the-art hashing frameworks on three benchmark datasets, \ie, CIFAR-10, NUS-Wide, and MS COCO.
\end{itemize}

%------------------------------------------------------------------------- 
\section{Related Works}\label{sec:relw}
%Hashing is gaining more attention due to its low storage requirement and fast retrieval speed. Supervised methods can achieve better performance compared to the unsupervised methods \cite{dpsh,csq,Wang2018SupervisedDH,Huang2019AccelerateLO}, but their usage is limited due to the large amount of annotation information required. Therefore, this paper focuses on unsupervised hashing methods. 
In this section, we will briefly introduce some unsupervised hashing methods here.

\noindent\textbf{Unsupervised Hashing.}\quad 
Early unsupervised hashing methods mainly focus on projecting images to compact representations by constraining the learned hash codes to fit several principles, \eg, quantization \cite{itq}, balancing \cite{deepbit}. Several recent works using deep learning pay attention to how to generate high quality hash codes \cite{mbe,greedyhash,cimon,hashgan}. Some others try to preserve the semantic similarity in hash codes \cite{sadh,distill,tbh}, while the majority of methods adopt image pseudo labels with pre-trained networks to convert the unsupervised hashing to fully a supervised learning \cite{zhang2017unsupervised}. Their performance is usually evaluated against ranked candidates. However, they did not try to sort them during training to mine their similarity.

\noindent\textbf{Hashing with Contrastive learning.}\quad Contrastive learning has been a very successful approach for unsupervised hashing tasks. Typical examples include CIMON \cite{cimon}, CIB \cite{cibhash}, NSH \cite{nsh}. All of these methods utilize the contrastive learning framework. As we mentioned before, these methods do not well combine the contrastive learning framework with the hash retrieval task. For example, both CIMON and CIB define the data-enhanced version of an image as a positive sample, and a negative sample is formed by sampling the views of different images. It leads to the possibility that images considered as negative samples may contain positive samples, which will have an impact on the retrieval results. On the other hand, although NSH considers this problem, they simply rank the similarity between anchor samples and select quantitative positive candidates, which does not take into account the similarity degree between the anchor images and augmented images.

\noindent\textbf{Mining Similarity for Unsupervised Hashing.}\quad Some methods based on mining similarity aim at solving unsupervised hashing tasks using pairwise methods, \eg, SSDH~\cite{ssdh} is a representative method studied in this area. It sets two thresholds at pairwise distances and constructs a similarity structure, and then image features are extracted and hash code learning is performed. However, using two rough thresholds to determine whether they are similar or not is usually unreliable. DistillHash~\cite{distill} extracts similarity signals using similarity signals from local structures, and further constructs an efficient and adaptive semantic graph, which is updated by decoding it in the context of an autoencoder for hash code learning. $\rm MLS^3RDUH$~\cite{MLS3RDUH} reconstructs a local semantic similarity structure by exploiting the intrinsic flow structure and cosine similarity in the feature space. DATE~\cite{date} improves the commonly used cosine distance by proposing a distribution-based metric. In contrast to these methods, WCH guides the learning of hash codes based on the weighted similarity between patches assigned to each anchor and the rest of the samples.

\begin{figure}[!t]
	\begin{center}
		\includegraphics[width=\linewidth]{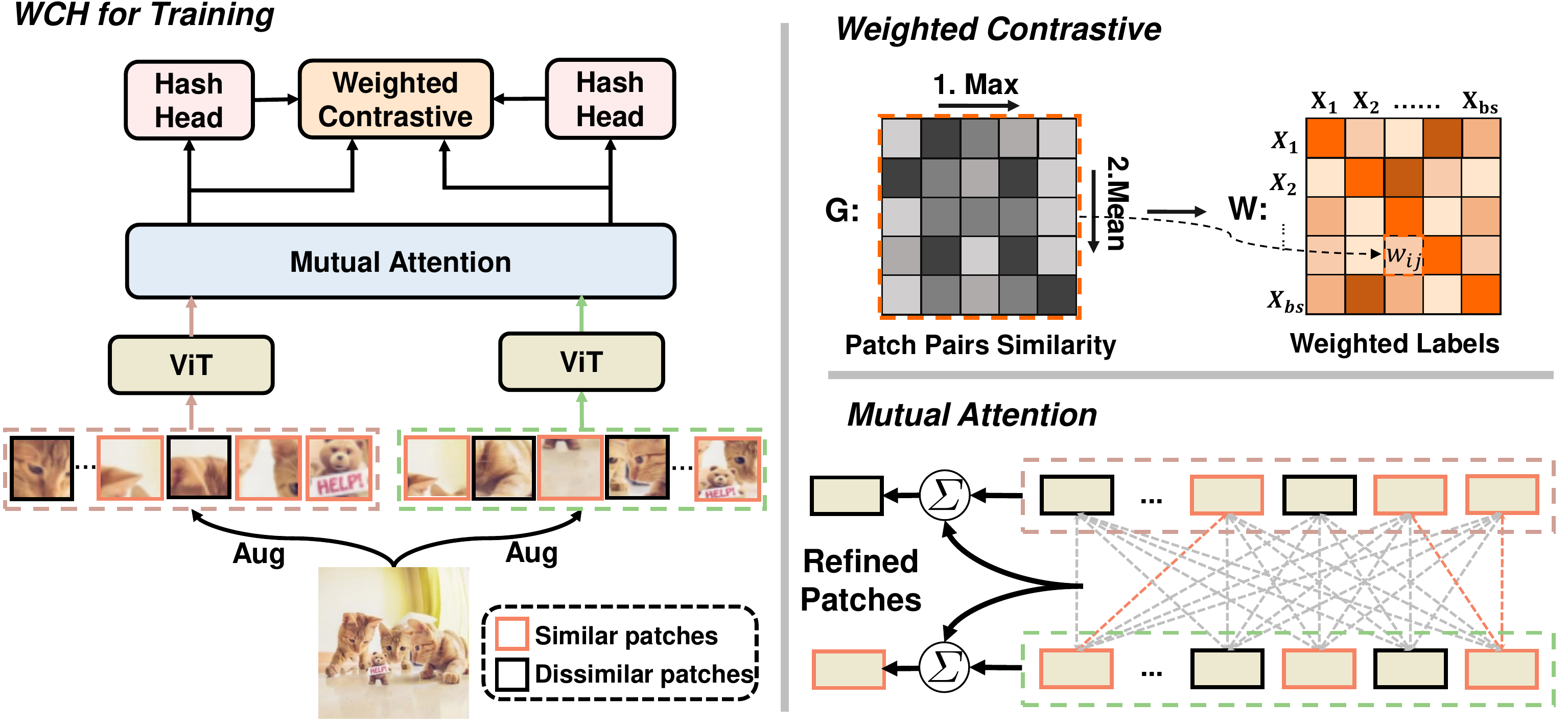}
	\end{center}
	\caption{Overall architecture of the proposed weighted contrastive hashing.}
 \label{fig_2}
\end{figure}

%------------------------------------------------------------------------- 
\section{Weighted Contrastive Learning} \label{sec:med}
\subsection{Preliminaries} \label{sec_vit}
To better explain our method in the next section, we first introduce some concepts and preliminary knowledge here.

\noindent\textbf{Patch Generation.} Following ViT \cite{vit}, we divide an image $\XX \in \mathbb{R}^{s \times s \times c}$ into non-overlapping patches $\xx_i \in \mathbb{R}^{p \times p \times c}$, where $i = 1, 2, \cdots, n$. It is obvious that $ s^2 = n \times p^2$ and $c$ is the number of image channels. 

\noindent\textbf{WCH encoder.} Recent work proposes the use of the ViT model as a universal feature extractor \cite{vithash}. Inspired by these works, we also use ViT as an encoder for our model. We first flatten the patches $\xx_{i}$ into a vector $\pp_i\in\mathbb{R}^{1 \times d}$, where $d$ is the dimension of the vector, and then use a trainable linear projector $\lp$ to map the vector to embedding. The output of this projection is referred to the patch embedding as follows:
\begin{equation}\label{eq_1}
    \EE_{i}=\lp (\pp_i),
\end{equation}
where $\EE_{i}$ is the patch embedding associated with the $i$-th patch. Unlike the standard ViT, our model does not use the class token. We add the position embedding into the patch embeddings, and the final embedding for ViT Input is:
\begin{equation}\label{eq_2}
    \mathbf{PE}_{i}=\EE_{i} + \mathbf{PoE},
\end{equation}
where, $\mathbf{PoE}$ stands for the position embedding, and $\mathbf{PE}$ is the final projecting embedding, which will be fed into the transformer encoder $f_{\theta}(\cdot)$.

%-------------------------------------------------------------------------
\noindent\textbf{Binarization.} In WCH, we revisit the problem of how to evaluate whether a candidate sample is comparable to an anchor in the contrastive learning framework and obtain higher quality hash codes in the image retrieval task. As for an image retrieval task, the goal is to learn a binary vector $\bb_i \in\{-1,1\}^{l}$ by mapping the data $\mathbf{x}_i$ into the encoder, where $l$ is the length of the hash code. In general, the hash code is obtained by the sign function:
\begin{equation}\label{eq_3}
    \mathbf{b}_i=\operatorname{sign}(h(\mathbf{x}_i))\in\{-1,1\}^{l},
\end{equation}
where $h(\cdot)$ is the encoding function, which mainly consists of the WCH encoder and a one-layer projector. Since the $\operatorname{sign}(\cdot)$ function is non-differentiable, we adopt a straight-through estimator (STE) \cite{bengio2013estimating} that allows back-propagation through $\mathbf{b}_i$.

\subsection{Overall Architecture}
WCH employs contrastive learning as an unsupervised framework, which typically defines positive pairs as different augmented parts of the same image and negative pairs as samples of different images. Given a batch of $N$ samples $\left\{\mathbf{X}^{1}, \cdots, \mathbf{X}^{i}, \cdots, \mathbf{X}^{N}\right\}$, it first goes through two different data augmenters to get two different views $\tilde{\mathbf{X}}^i$ and $\hat{\mathbf{X}}^i$. Then, we divide each image into $n$ non-overlapping patches, $\tilde{\mathbf{X}}^i=\left[\tilde{\boldsymbol{x}}^i_{1}; \cdots; \tilde{\boldsymbol{x}}^i_{k}; \cdots; \tilde{\boldsymbol{x}}^i_{n}\right], \hat{\mathbf{X}}^i=\left[\hat{\boldsymbol{x}}^i_{1}; \cdots; \hat{\boldsymbol{x}}^i_{k}; \cdots; \hat{\boldsymbol{x}}^i_{n}\right]$. As we described in Sec.~\ref{sec_vit}, we employ ViT as the encoder, and feed the patches into it to generate the corresponding encoded features $f_{\theta}(\tilde{\boldsymbol{x}}^i_k )$ and $f_{\theta}(\hat{\boldsymbol{x}}^i_k )$. 

During traditional augmentation, the augmented two images are usually taken as a similar pair to guide the training direction. However, some cropped images containing background only are totally different from others, which might damage the training process. Therefore, we employ the Mutual Attention (MA) module to re-weight the image patches to guarantee similarity between them. After that, the weighted image similarity is calculated by computing the patch similarity between different images, and it is subsequently used to construct the final weighted contrastive loss function. The overall architecture is illustrated in Fig. \ref{fig_2}.

\subsection{Mutual Attention}

Given any two encoded patches $f_{\theta}(\tilde{\boldsymbol{x}}^i_k )$ and $f_{\theta}(\hat{\boldsymbol{x}}^i_t )$ in the corresponding augmented pair, the similarity of them can be calculated as
\begin{equation}
    \mathbf{s}_{k,t}=f_{\theta}(\tilde{\boldsymbol{x}}^i_k )^T f_{\theta}(\hat{\boldsymbol{x}}^i_t ).
\end{equation}
Therefore, we can construct a similarity matrix $\mathbf{S} \in \mathbb{R}^{n \times n}$, which measures the similarity between each patch of the augmented pair. Then, we normalize the row vectors and column vectors respectively with softmax function:
\begin{equation}
\left\{  
\begin{array}{lr}  
    \mathbf{S}^1 &= softmax(\mathbf{s}_{1\ast}, \cdots, \mathbf{s}_{i\ast}, \cdots, \mathbf{s}_{n\ast}) \\
    \mathbf{S}^2 &= softmax(\mathbf{s}_{\ast 1}, \cdots, \mathbf{s}_{\ast j}, \cdots, \mathbf{s}_{\ast n})
\end{array}  
\right. ,
\end{equation}
where $\mathbf{s}_{i\ast}$ stands for the $i$-th row of $\mathbf{S}$, and $\mathbf{s}_{\ast j}$ means the $j$-th column of $\mathbf{S}$. Then, the refined patch vector is reconstructed with the following calculation:
\begin{equation}\label{eq_5}
\left\{  
\begin{array}{lr}
    \tilde{\boldsymbol{f}}^i_k &= \sum_{j=1}^{n}s^2_{j,k}f_{\theta}(\tilde{\boldsymbol{x}}^i_j ) \\
     \hat{\boldsymbol{f}}^i_k  &= \sum_{j=1}^{n}s^1_{k,j}f_{\theta}(\hat{\boldsymbol{x}}^i_j )
\end{array}  
\right.,
\end{equation}
where $s^1_{j,k}$ is the $j$-th row and $k$-column element of $\mathbf{S}^1$, and $s^2_{k,j}$ is analogously defined. After this operation, the refined augmented pair can be guaranteed to be similar, and thence be undoubtedly treated as a positive pair.

\subsection{Weighted Similarities Calculation}\label{sec_43}

%Unsupervised image retrieval tasks based on contrastive learning have shown significant performance. However, since only the augmented parts of the same image are considered to be from the same class, this solution ignores the semantic content of the other images. Besides, this also leads to conflicting objectives and hinders the convergence of the objective function based on contrastive learning. 

Previous works often use embedding vectors to explore the relationship between different images. Specifically, most existing unsupervised hashing methods assume binary similarity between two images, \ie, two images can be similar (positive sample) or dissimilar (negative sample). For example, NSH \cite{nsh} uses hash codes to calculate the degree of similarity between images, and then ranks them according to the magnitude of similarity, and selects the top $m$ positive samples according to the result of the ranking, \ie, determines that these $m$ samples and anchor are similar. However, selecting positive samples based on similarity like this will cause two problems. First, there may be noise in the positive samples. Since the number of positive samples is set to a fixed value, forcing a fixed number of positive samples based on the ranking results will slow down the convergence of the model. 
Second, the results of the first $m$ closest samples may not be equivalent. For an anchor image $\XX_i$, $\XX_j$ is one of the selected $m$ positive samples that are similar to $\XX_i$ in one iteration. However, it is possible that in another iteration, $\XX_j$ is not one of the $m$ closest samples since there are more similar images in this training batch. In this situation, $\XX_j$ will be treated as a negative sample of $\XX_i$, which results in inconsistency with the former one, and this conflict will damage the training process and cause the training to fail to converge.

\noindent\textbf{Weighted Labels Processing.}  To tackle these problems, instead of adopting a strategy such as selecting positive samples, we reformulate the rules for computing the similarity between images and use the obtained similarity to re-weight the contrastive loss to capture the semantic information that may overlap between the anchor and negative samples. In WCH, we exploit the fine-grained interaction results between patches to explore the relationship between different images.

Specifically, suppose there are two patch features $\boldsymbol{f}^{i}_k$ and $\boldsymbol{f}^{j}_t$ extracted from two different images $\XX^i$ and $\XX^j$, respectively. The similarity between them can be defined as
\begin{equation}\label{eq_6}
g^{ij}_{kt}=(\boldsymbol{f}^{i}_k)^T \boldsymbol{f}^{j}_t.
\end{equation}

Therefore, the similarity matrix of $\XX^i$ and $\XX^j$ can be constructed as $\mathbf{G}^{ij} \in \mathbb{R}^{n \times n}$. For each row in $\mathbf{G}^{ij}$, the max value represents the most similar batches in $\XX^i$ and all patches in $\XX^j$, and the mean of the max values of each row is the similarity of $\XX^i$ and $\XX^j$:
\begin{equation} \label{eq_8}
    \begin{aligned}
        w_{ij} = \operatorname{mean}(\underset{row}{\maxx}(\mathbf{G}^{ij})),
    \end{aligned}
\end{equation}
where $\underset{row}{max}(\cdot)$ means to take the maximum value according to the row direction, and $mean(\cdot)$ stands for calculating the mean value of the vector. For a mini-batch containing $bs$ images, including one augmented image and $bs-1$ other images, the similarity matrix $\mathbf{W} \in \mathbb{R}^{bs \times bs}$ can be constructed with Eq. \ref{eq_8}. To fit the value of $\mathbf{W}$ within a proper range, we conduct a temperature weighted row softmax as $\mathbf{W}_{i\ast} = softmax(\mathbf{W}_{i\ast}/\tau_w)$, where $\tau_w$~\cite{temperature} is the temperature coefficient. Furthermore, to guarantee that the augmented images are equivalent to the anchor images, we divide each row with the element $w_{ii}$:
\begin{equation}\label{eq_9}
    \begin{aligned}
        \mathbf{W} = \diag(\diag(\mathbf{W}))^{-1}\mathbf{W},
    \end{aligned}
\end{equation}
where $\diag(\cdot)$ means extracting the diagonal vector from a matrix or constructing a diagonal matrix with a vector.

\subsection{Training and Inference}
For training, we use the maximum similarity between patches to guide the contrastive objective~\cite{infonce}:
\begin{equation}\label{eq_10}
\mathcal{L}_{\mathrm{WCE}} =-\sum_{i=1}^{bs} \sum_{j=1}^{bs} w_{ij} \log \frac{\exp \left(\tilde{\mathbf{b}_{i}} \hat{\mathbf{b}}_{j}^\intercal / l / \tau\right)}{\sum_{k=1}^{bs} \exp \left(\tilde{\mathbf{b}_{i}} \hat{\mathbf{b}}_{k}^\intercal / l / \tau\right)},
\end{equation}
where $\tau$ is the temperature scale. Finally, the loss function is formulated as
\begin{equation}\label{eq_11}
\mathcal{L}_{\mathrm{WCH}} =\mathcal{L}_{\mathrm{WCE}}+ {\mathcal{L}_{R}},
\end{equation}
where $\mathcal{L}_{R}$ refers to the quantization loss and bit balancing loss~\cite{dh}. The whole learning procedure is shown in Alg.~\ref{algorithm}.

\noindent\textbf{Inference process.} In the inference process, WCH abandons the MA and weighted labeling modules dedicated to training and keeps only the encoder and hash head for generating hash codes characterizing the semantic information of the images. The Hamming distance between the hash codes of the images is then computed to accomplish the retrieval task.

\begin{algorithm}[t]
    \label{algorithm}
	\small
	\caption{The Training Procedure of WCH.}
	\label{alg}
	\textbf{Input:}\hspace{0mm} Dataset $\mathcal{D}=\{\xx_i\}_{i=1}^N$ and batch size $n$.\\
	\textbf{Output:}\hspace{0mm} Network parameters $\theta$.\\
	%Randomly initialize $\mathbf{H}\in\{-1,1\}^{M\times N}$\\
	\For{$\mathtt{batch~in}~\mathcal{D}.repeat()$}{
	    $\variable{batch_1},~\variable{batch_2}=\func{aug}(\variable{batch}),~\func{aug}(\variable{batch})$\\
	    $\variable{f_1,~f_2} = \func{M_\theta}(\variable{batch_1}),~\func{M_\theta}(\variable{batch_2})$\\
	    \textcolor[rgb]{0.5,0.5,0.5}{\#~mutual attention}\\
	    $\variable{sim}=\func{einsum}(\param{'nid,njd\rightarrow nij'},~\variable{f_1,~f_2})$\\
	    $\variable{f_1}=\func{einsum}(\param{'nid,ndj\rightarrow nij'},~\variable{\func{softmax}(\variable{sim.T}),~f_1})$\\
	    $\variable{f_2}=\func{einsum}(\param{'nid,ndj\rightarrow nij'},~\variable{\func{softmax}(\variable{sim}),~f_2})$\\
	    \textcolor[rgb]{0.5,0.5,0.5}{\#~weighted similarities calculation}\\
	    $\variable{sim}=\func{einsum}(\param{'nid,mjd\rightarrow nmij'},~\variable{f_1,~f_2})$\\
	    $\variable{sim}=\func{softmax}(\variable{sim}.\func{max}(\param{-1}).\func{mean}(\param{-1})/\variable{\tau_{w}})$\\
	    $\variable{weighted}=\func{matmul}(\func{diag}(\func{diag}(\variable{sim}))^{-1},~\variable{sim})$\\
	    \textcolor[rgb]{0.5,0.5,0.5}{\#~hashing}\\
	    $\variable{b_1,~b_2} = \func{hash}(\variable{f_1.\func{mean}(\param{1})}),~\func{hash}(\variable{f_2.\func{mean}(\param{1})})$\\
	    $\variable{logits}=\func{softmax}(\func{matmul}(\variable{b1,b2.T})/l/\variable{\tau})$\\
	   \textcolor[rgb]{0.5,0.5,0.5}{ \#~weighted corss entropy}\\
	    $\variable{loss}{=}~\func{cross\_entropy}(\variable{logits,~weighted})$\\
	    $\variable{loss}.\func{backward}()$
	}
\end{algorithm}

\section{Discussion}
\noindent\textbf{Remark 1: Why Do We Choose the ViT Encoder?}~In WCH, our key idea is to use the patch-level semantic information captured by ViT~\cite{vit} as a benchmark to measure the degree of similarity between arbitrary images and assign corresponding weights to each pair of images by aggregating the similarity between patches to measure the degree of similarity. Unlike the recent self-supervised visual representation learning-based approaches~\cite{cibhash,nsh}, they only determine similar samples by the global feature similarity of the whole image. Instead, we introduce a novel inter-patch-based fine-grained interaction module using the ViT model, enabling fine-grained interactions between patches and each pair of images to mine more detailed semantic alignment. 

Furthermore, we use the ViT model to address the problems posed by contrastive learning methods that rely on instance discrimination tasks. As mentioned before, positive native pairs are defined as different views of the same image, while negative pairs are formed by sampling views of different images. This common approach ignores their semantic content. Our approach, on the other hand, fully exploits the semantic content of the images and makes reasonable use of fine-grained interaction results as a measure of similarity between images, as detailed in Sec.~\ref{sec_43}.

\noindent\textbf{Remark 2: Why Mutual Attention Helps?} ~First, note the phenomenon that most models construct positive and negative samples by treating the same images produced by different augmenters as positive pairs, while the rest of the samples are considered as negative pairs. However, this manually designed approach involves many manual choices, and inappropriate data augmentation schemes may severely alter the image structure, resulting in data-enhanced images that do not possess label-preserving properties, \ie, images undergo transformations that may lose high-level semantic information. For example, a common data augmentation scheme is random cropping, which may randomly crop out the sample information that contains label-related information for single-labeled images. Similarly, for multi-labeled images, where the position and size of objects vary greatly, the random cropping method will most likely crop out some objects in multi-labeled images, making the sample information contained in multi-labeled images reduced. This operation will lead to asymmetric semantic information between the anchor and the positive samples.

The mutual attention module in Fig.~\ref{fig_2} reconstructs the feature vectors associated with the pictures based on the similarity between positive sample pairs of patches. Therefore, it can be seen as a specific attention mechanism. Intuitively, it focuses our attention on the degree of similarity of patch pairs. The attention fraction is used so that the feature vectors of each patch carry information about other patches to different degrees. More specifically, this attention mechanism is very useful for multiple patches, especially when there are many classes of objects and the positions are highly variable.

\noindent\textbf{Remark 3: Why Do We Gather $\mathbf{W}$ in This Way?} ~The purpose of Weighted Labels is to use the maximum similarity between patches to guide the contrastive objective. Using the maximum similarity between patches in Eq.~\ref{eq_8}, we can get the most similar patch pair among all patches in the two images. Then the sum of the maximum similarity is averaged. The model learns the fine-grained semantic alignment between patches by applying the weighted label to the contrastive loss.

%-------------------------------------------------------------------------

\section{Experiments}\label{sec:exp}
In this section, we conduct experiments on three datasets, including one single-labeled dataset and two multi-labeled datasets, to evaluate our method.

\begin{table*}[t]
	\begin{center}
		\caption{Performance comparison (mAP) of WCH and the state-of-the-art \textbf{unsupervised} hashing methods. *Note that we use a more common setting on NUS-WIDE with the 21 most frequent classes, while some papers report results on 10 classes.}
  \label{tab_map}
	\small
	\resizebox{\textwidth}{!}{
		\begin{tabular}{l l ccc  ccc  ccc}
			\hline
			\multirow{2}{*}{\textbf{Method}}&\multirow{2}{*}{\textbf{Reference}}&\multicolumn{3}{c}{\textbf{ CIFAR-10}}&\multicolumn{3}{c}{\textbf{NUS-WIDE} }&\multicolumn{3}{c}{\textbf{MS COCO}}\\\cline{3-11}
			%\rowcolor{gray!15}
			&& 16 bits& 32 bits & 64 bits& 16 bits& 32 bits & 64 bits& 16 bits& 32 bits & 64 bits\\ \hline\hline
			%LSH~\cite{lsh}&STOC02& 0.106& 0.102 & 0.105 & 0.239& 0.266 & 0.266& 0.353& 0.372& 0.341\\%& 0.152 & 0.141 & 0.165\\
			%SpH~\cite{sh}&NeurIPS09& 0.272& 0.285& 0.300 & 0.517& 0.511& 0.510 & 0.527& 0.529& 0.546\\%& 0.185& 0.271& 0.350\\
			AGH~\cite{agh}&ICML11& 0.333& 0.357 & 0.358 & 0.592& 0.615& 0.616&0.596 &0.625 & 0.631\\%& 0.241& 0.326& 0.379\\
			%SpherH~\cite{sph}&CVPR12& 0.254 & 0.291 & 0.333 & 0.495 & 0.558 & 0.582 & 0.516& 0.547 &0.589\\%& 0.110 & 0.187& 0.259\\
			%KMH~\cite{he2013k}&CVPR13& 0.279 & 0.296& 0.334 & 0.562& 0.597 & 0.600 & 0.543 & 0.554 & 0.592\\% & 0.202 & 0.297 & 0.390\\
			ITQ~\cite{itq}&PAMI13& 0.305& 0.325& 0.349& 0.627& 0.645& 0.664& 0.598& 0.624& 0.648\\%& 0.217& 0.317& 0.391\\
			DGH~\cite{dgh}&NeurIPS14& 0.335& 0.353& 0.361& 0.572& 0.607& 0.627& 0.613& 0.631&0.638\\\hline% &0.270 &0.348 &0.373\\\cline{1-13}
			%DeepBit~\cite{deepbit}&CVPR16& 0.194& 0.249& 0.277& 0.392& 0.403& 0.429& 0.407& 0.419 & 0.430 \\%& 0.204& 0.283 &0.286\\
			SGH~\cite{sgh}&ICML17 & 0.435& 0.437 & 0.433& 0.593& 0.590& 0.607& 0.594& 0.610& 0.618\\%& 0.447& 0.500& 0.523\\
			BGAN~\cite{bgan}&AAAI18& 0.525& 0.531& 0.562& 0.684& 0.714& 0.730& 0.645& 0.682& 0.707\\%& 0.499& 0.535& 0.574\\
			BinGAN~\cite{bingan}&NeurIPS18& 0.476& 0.512& 0.520& 0.654& 0.709& 0.713& 0.651& 0.673& 0.696\\%& 0.505& 0.584& 0.612\\
			GreedyHash~\cite{greedyhash}&NeurIPS18& 0.448& 0.473& 0.501& 0.633& 0.691& 0.731 & 0.582& 0.668& 0.710\\%& 0.186& 0.578& 0.558\\
			HashGAN~\cite{hashgan}&CVPR18& 0.447& 0.463& 0.481& -& -& -& -& -& -\\%& -& -& -\\
			DVB~\cite{dvbj}&IJCV19& 0.403& 0.422& 0.446& 0.604& 0.632& 0.665& 0.570& 0.629& 0.623\\%& 0.398& 0.452& 0.465\\
			DistillHash~\cite{distill}& CVPR19& 0.284& 0.285& 0.288& 0.667& 0.675& 0.677& -& -& -\\%\hline\hline%& -& -& -\\\hline\hline
			{TBH}~\cite{tbh}& CVPR20&{0.532}&{0.573}&{0.578}&{0.717}&{0.725}&{0.735}&{0.706}&{0.735}&{0.722}\\%&\textbf{0.560}&\textbf{0.619}&\textbf{0.6}
			{$\rm MLS^{3}RDUH$}~\cite{MLS3RDUH}& IJCAI20&{0.369}&{0.394}&{0.412}&{0.713}&{0.727}&{0.750}&{0.607}&{0.622}&{0.641}\\%&\textbf{0.560}&\textbf{0.619}&\textbf{0.626}\\\hline
			{DATE}~\cite{date}& MM21&{0.577}&{0.629}&{0.647}&{0.793}&{0.809}&{0.815}& -& -&-\\%&\textbf{0.560}&\textbf{0.619}&\textbf{0.626}\\\hline
			{MBE}~\cite{mbe}& AAAI21&{0.561}&{0.576}&{0.595}&{0.651}&{0.663}&{0.673}& -& -&-\\%&\textbf{0.560}&\textbf{0.619}&\textbf{0.626}\\\hline
			{CIMON}~\cite{cimon}*& IJCAI21&{0.451}&{0.472}&{0.494}& -& -&-&-&-&-\\%&\textbf{0.560}&\textbf{0.619}&\textbf{0.626}\\\hline
			{CIBHash}~\cite{cibhash}& IJCAI21&{0.590}&{0.622}&{0.641}&{0.790}&{0.807}&{0.815}&{0.737}&{0.760}&{0.775}\\%&\textbf{0.560}&\textbf{0.619}&\textbf{0.626}\\\hline
			{SPQ}~\cite{spq}& ICCV21&{0.768}&{0.793}&{0.812}&{0.766}&{0.774}&{0.785}& -& -&-\\%&\textbf{0.560}&\textbf{0.619}&\textbf{0.626}\\\hline
%			{MeCoQ}~\cite{mecoq}& AAAI22&{0.682}&{0.697}&{0.711}&{0.802}&{0.822}&{0.832}& -& -&-\\%&\textbf{0.560}&\textbf{0.619}&\textbf{0.626}\\\hline 
			{NSH}~\cite{nsh}& IJCAI22&{0.706}&{0.733}&{0.756}&{0.758}&{0.811}&{0.824}&{0.746}&{0.774}&{0.783}\\\hline%&\textbf{0.560}&\textbf{0.619}&\textbf{0.626}\\\hline
			\textbf{WCH}& \textbf{Proposed}&\textbf{0.897}& \textbf{0.910}& \textbf{0.932}& \textbf{0.799}& \textbf{0.823}&\textbf{0.838}&\textbf{0.776}&\textbf{0.808}&\textbf{0.834}\\\hline
		\end{tabular}
	}
\end{center}
\end{table*}

\subsection{Datasets and Evaluation Metrics}
Three benchmark datasets are used in our experiments. \textbf{CIFAR-10}~\cite{cifar} consists of 60,000 images from 10 classes. We follow the common setting \cite{hashgan} and select 10,000 images (1000 per class) as the query set. The remaining 50,000 images are regarded as the database.
\textbf{NUS-WIDE}~\cite{nus} has of 81 categories of images. We adopt the 21-class subset following \cite{nsh}. 100 images of each class are utilized as a query set, with the remaining being the gallery.
\textbf{MS COCO}~\cite{coco} 
is a benchmark for multiple tasks. We use the conventional set with 12,2218 images. We randomly select 5,000 images as queries with the remaining ones the database.

\textbf{Evaluation Metric.} To compare the proposed method with the baselines, we adopt several widely-used evaluation metrics, including the mean average precision (mAP), top-K precision (P@K), precision-recall (PR) curves \cite{zhang2017unsupervised}. %Specifically, the first one is calculated according to the Hamming ranking, while the last two are based on hash lookup. We use all the returned results to calculate mAP and set K to 1000 for P@K.

 \subsection{Implementation Details}
For all three datasets, the images were resized to $224 \times 224 \times 3$ and we adopt the image augmentation strategies of~\cite{simclr}. The standard ViT-Base~\cite{vit} was used as the backbone, with patches of size and number 16 and 196, respectively. As in previous work~\cite{cimon,cibhash}, we loaded a pre-trained model trained on ImageNet to accelerate the convergence. We used the cosine decay method and trained 50 epochs for all models, with the initial learning rate set to $1 \times 10^{-5}$.

\subsection{Comparison with the SotA}
\noindent\textbf{Baselines.} We compare WCH against 18 state-of-the-art baselines, including 3 traditional unsupervised hashing methods and 15 recent unsupervised hashing methods. For fair comparisons, all the methods are reported with identical training and test sets. Additionally, the shallow methods are evaluated with the same deep features as the ones we are using.
 
\begin{figure}[!t]
	\begin{center}
		\includegraphics[width=0.85\linewidth]{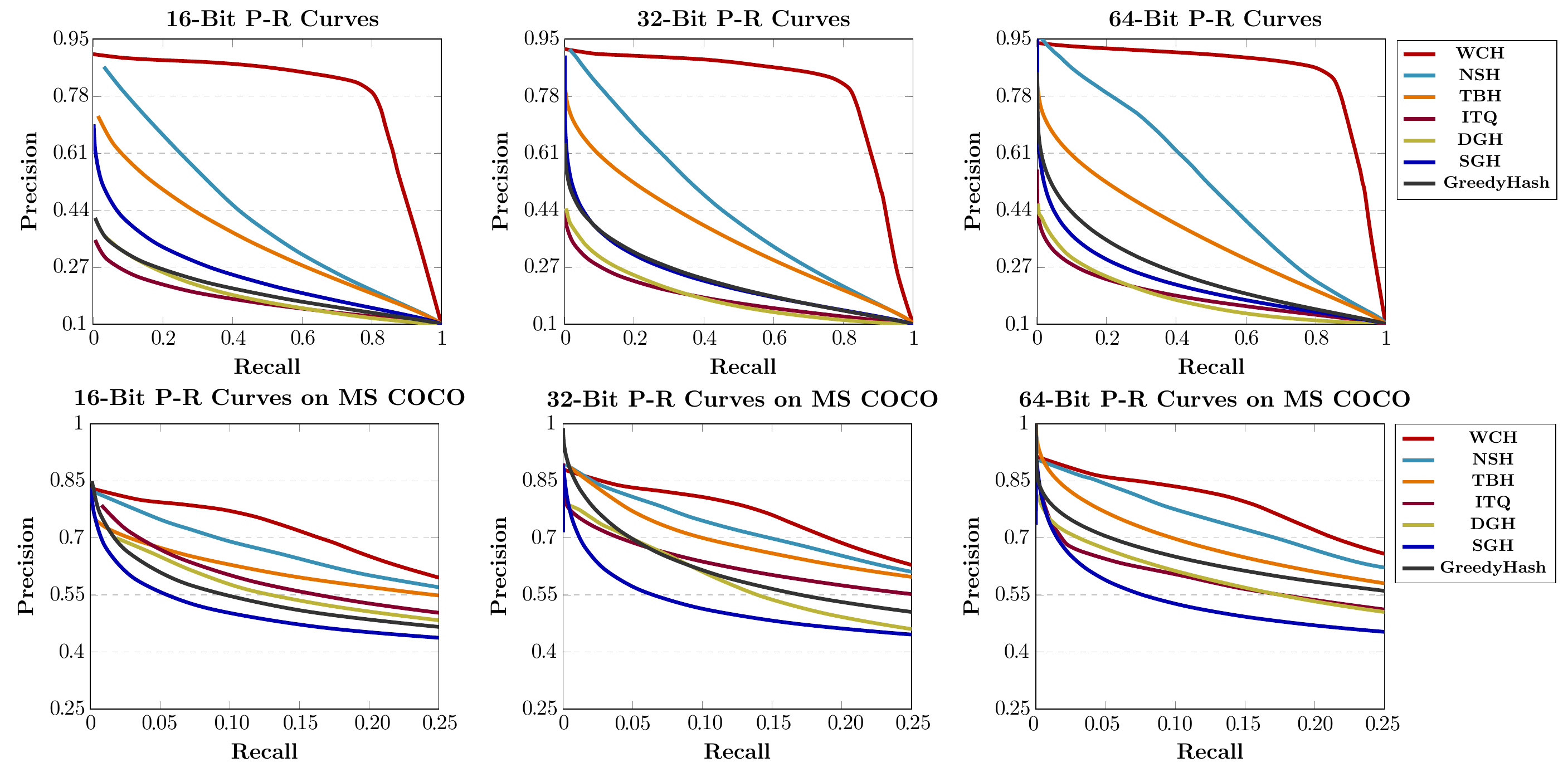}
	\end{center}
	\caption{P-R curves comparison with other methods on CIFAR-10 and MS COCO.}
\label{figure3}
\end{figure}

\noindent\textbf{Results.}~ Tab.~\ref{tab_map} shows the retrieval performance in mAP and Tab. \ref{tab_pre} demonstrates the precision of the first 1000 returned images. It can be clearly observed that WCH obtains the best results on all three datasets for the two metrics. Another interesting observation is that WCH significantly outperforms the previous works CIBHash and NSH on different hash bits and datasets. Note that all three methods use contrastive learning. In addition, the P-R curves of WCH and several baselines on CIFAR-10 and MS COCO are reported in Fig. \ref{figure3}, from which it can also be discovered that the curves of our method are highly above those of other methods for all three different code lengths.

\begin{table}[!t]
\caption{P@1000 results on CIFAR-10 and MS COCO.}
	\centering
		\begin{tabular}{l ccc ccc}
			\hline
			\multirow{2}{*}{\textbf{Method}} &\multicolumn{3}{c}{\textbf{ CIFAR-10}}&\multicolumn{3}{c}{\textbf{MS COCO}}\\\cline{2-7}
			&16 bits& 32 bits & 64 bits& 16 bits& 32 bits & 64 bits\\\hline\hline
			%			KMH & 0.242& 0.252& 0.284& 0.557 & 0.572& 0.612 \\
			%			SpherH & 0.228& 0.256& 0.291& 0.525 & 0.571 & 0.612\\
			%			%LSH~\cite{lsh} & 9.95& 9.65& 10.02&0.99&0.92&1.09\\
			%ITQ & 0.276& 0.292 & 0.309 & 0.607& 0.637& 0.662\\
			%			SpH & 0.238 & 0.239& 0.245& 0.541& 0.548& 0.567\\
			AGH & 0.306& 0.321& 0.317& 0.602& 0.635& 0.644\\
			DGH & 0.315& 0.323& 0.324 & 0.623& 0.642& 0.650\\
			HashGAN & 0.418& 0.436& 0.455 &-&-&-\\
			%SGH & 0.387 & 0.380 & 0.367 & 0.604 & 0.615 & 0.637\\
			GreedyHash & 0.322 & 0.403 & 0.444 & 0.603 & 0.624 & 0.675 \\
			{TBH} & {0.497}& {0.524}& {0.529}& {0.646}& {0.698}& {0.701}\\
			CIBHash& 0.526& 0.570& 0.583& 0.734& 0.767& 0.785\\
			NSH& 0.691& 0.716& 0.744& 0.733& 0.770& 0.805\\
			\hline
			\textbf{WCH}&\textbf{0.889}&\textbf{0.902}&\textbf{0.923}&\textbf{0.795}& \textbf{0.830}& \textbf{0.855}\\
			\hline
		\end{tabular}
\label{tab_pre}
\end{table}

\subsection{Ablation Studies}  
In this subsection, we considered the following ablation experiments to verify the effectiveness and contribution of each component of WCH, and the specific results are shown in Tab.~\ref{tab_abl}.
\begin{enumerate}[label=\color{black}(\roman*),wide,labelindent=0pt]
\item\textbf{ViT Baseline.}\label{bl_1} We first investigate the enhancements that the ViT backbone brings to the unsupervised hashing domain. In this baseline, the class token covering global features is applied directly to the hash head to generate a hash code characterizing the image. Subsequent contrastive loss is used to update the network parameters, which form a design close to the CIB~\cite{cibhash} except that the network backbone differs. Regrettably, the application of the ViT backbone alone is not sufficient to improve the performance of the unsupervised hashing.

\item\textbf{Without $\mathcal{L}_{R}$.}\label{bl_2} We also reveal the impact of traditional quantization loss and bit balance loss~\cite{dh} on WCH. It can be seen that these conventional regularizers have no significant improvement in the encoding quality. As a result, we can attribute the good performance entirely to our design.

\item\textbf{MA $\rightarrow$ mean.}\label{bl_3} We use this baseline to demonstrate the validity of our MA module. Here we remove the mutual attention mechanism of anchor and positive samples in Eq.~\ref{eq_5} and replace it with the averaging operation. Although it also achieves trivially good results, there is still a noticeable margin of difference with the performance of WCH, which indicates that the motivation of mutual attention can play a positive role.

\item\textbf{Weighted $\rightarrow$ hard.}\label{bl_4} This baseline does not use weighted labels, but rather the most fundamental hard labels, which means that the weighted contrastive learning degrades to the standard contrastive learning. The non-negligible performance degradation in Tab.~\ref{tab_abl} precisely illustrates the shortcoming of standard contrastive learning, which cannot close the distance between the anchor and similar negative samples in the feature space. This also highlights the crucial role played by our core motivation from the side.

\item\textbf{Without scale.}\label{bl_5} In this baseline, we remove the operation defined in Eq.~\ref{eq_9} and simply use the similarity matrix $\mathbf{W}$ in Eq.~\ref{eq_8} as a weighted label during the calculation of the loss. We can strikingly see an unexpectedly dramatic performance slippage. Hence, affine mapping based on positive sample similarity is a key factor to guarantee the effectiveness of weighted comparison learning.

\begin{table}[!t]
\caption{Ablation study results of mAP@1000 on MS COCO. The baselines are constructed by replacing some key modules of WCH.}
\label{tab_abl}
	\centering	
		\begin{tabular}{ll ccc}
			\hline
			&\textbf{Baseline}& \textbf{16 bits}& \textbf{32 bits}& \textbf{64 bits}\\\hline\hline
			\ref{bl_1}& ViT Baseline & 0.573& 0.595& 0.622\\
			%2&Swapped bottlenecks& 0.470 & 0.525 & 0.517\\
			\ref{bl_2}& Without $\mathcal{L}_{R}$& 0.773 & 0.810 & 0.828\\
			\ref{bl_3}& MA $\rightarrow$ $\rm mean$& 0.742& 0.782& 0.805\\
			\ref{bl_4}& weighted $\rightarrow$ $\rm hard$ & 0.738& 0.777& 0.799\\
			\ref{bl_5}& Without scale & 0.461& 0.479& 0.491 \\\hline
			&\textbf{WCH}& \textbf{0.776}& \textbf{0.808}& \textbf{0.834}\\\hline
		\end{tabular}
	\centering
\end{table}

 \begin{figure}[t]
	\centering
	\includegraphics[trim=0 0 0 0,clip,width=0.7\textwidth]{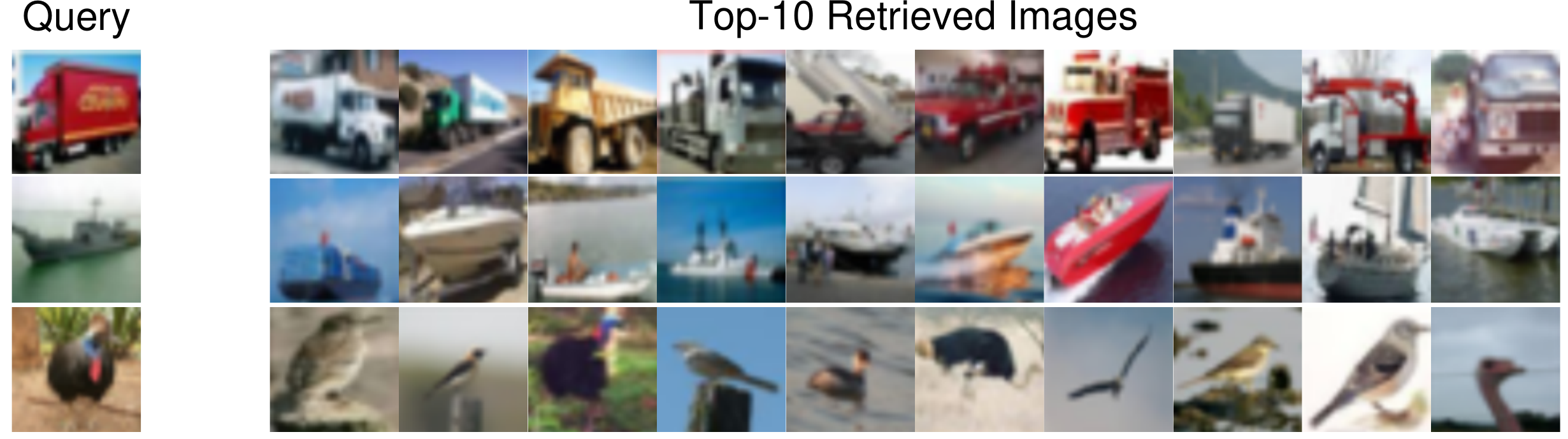}
	\caption{Examples of top-10 retrieved results of 32-bit on CIFAR-10.}
	\label{fig:top10}
\end{figure}

\begin{figure}[!t]
	\begin{center}
		\includegraphics[width=0.7\linewidth]{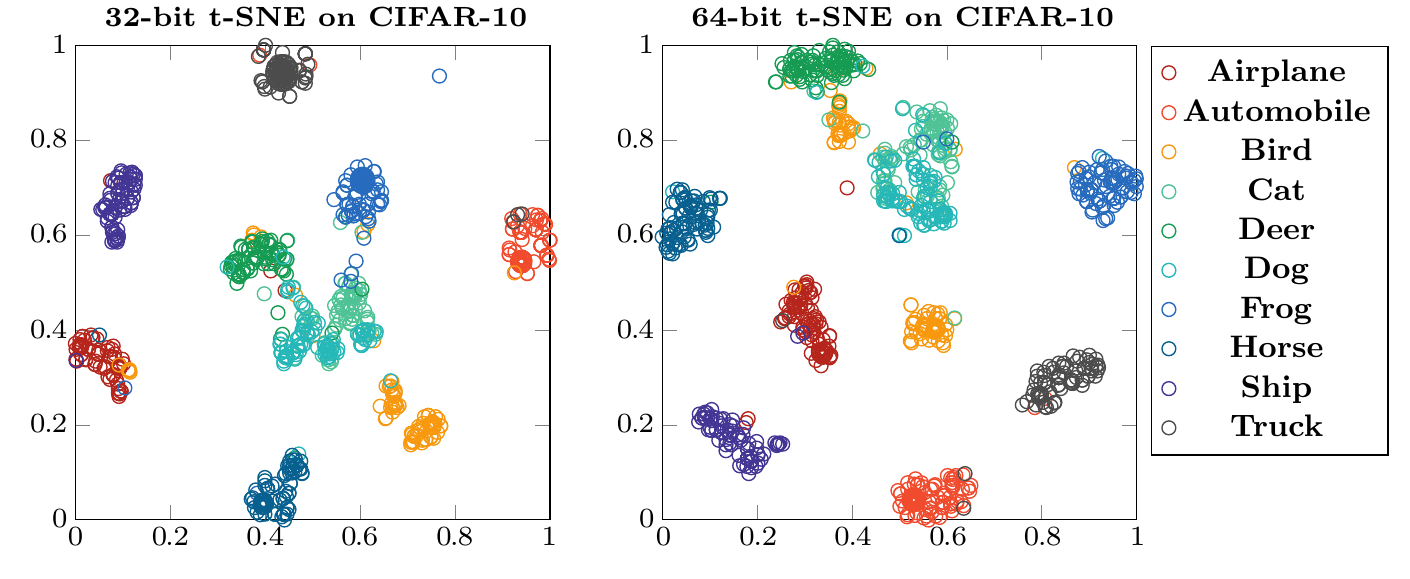}
	\end{center}
	\caption{32-bit and 64-bit t-SNE visualization results on CIFAR-10.}
 \label{fig_4}
\end{figure}

 \begin{figure}[t]
	\centering
	\includegraphics[trim=0 0 0 0,clip,width=0.95\textwidth]{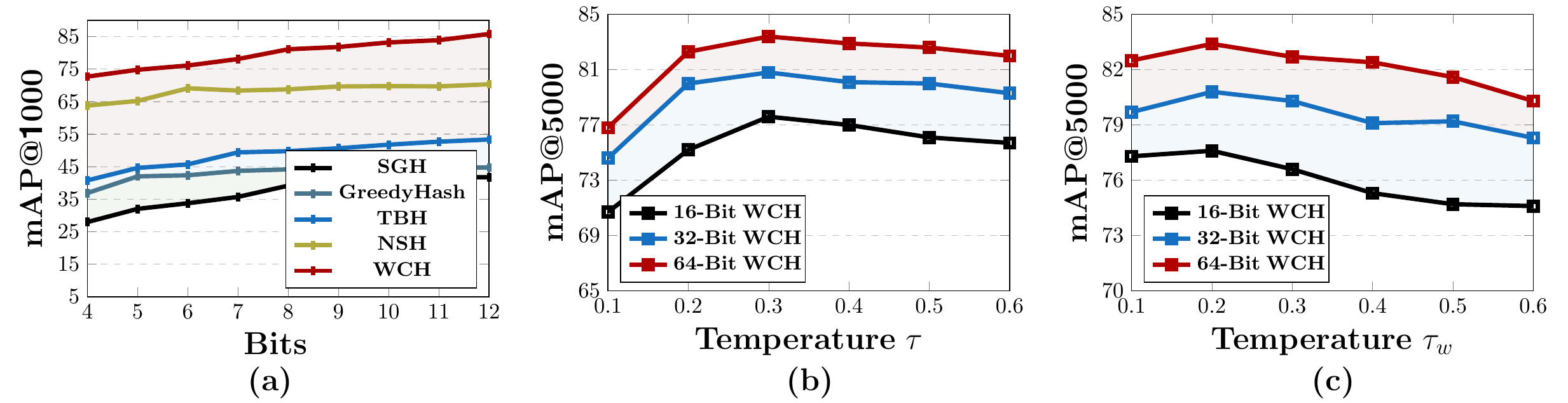}
	\caption{\textbf{(a)} mAP@1000 results with extremely short code lengths on CIFAR-10. \textbf{(b)}\&\textbf{(c)}Effects of different temperatures $\tau$ and $\tau_w$ on MS COCO.}
	\label{fig:para}
\end{figure}

\noindent\textbf{Results.} Baseline \ref{bl_1} contradicts our intuition that directly replacing the backbone network with ViT can not bring meaningful performance improvement. Baseline \ref{bl_3} shows that MA is an effective solution to deal with the information asymmetry problem for positive samples. We use baseline \ref{bl_4} to validate our core motivation that weighted contrast learning can substantially alleviate the class collision problem of negative samples and thus further improve the retrieval performance. %All these modules combined together can achieve optimal results for WCH.
\end{enumerate}

\subsection{Visualization and Hyper-parameters}
\noindent \textbf{Visualization.} To more intuitively demonstrate the performance of our method, we show the retrieved top-10 images on CIFAR-10 in Fig. \ref{fig:top10}, where a high semantic accuracy can be observed from the results. In addition, to show whether the embedded hash codes are discriminative enough for retrieval, the t-SNE plots \cite{tsne} of hash codes for both 32-bit and 64-bit on CIFAR-10 are also illustrated in Fig. \ref{fig_4}, where the plotted dots of different classes show obvious boundaries between them, which means that the generated codes are separable and shows the consistency with other results.

\noindent \textbf{Hyper-parameters.} In Fig.~\ref{fig:para}(a), we show the results for very short hash code lengths on CIFAR10. Although the performance varies slightly depending on the hyperparameter settings, it is generally stable and state-of-the-art. We also evaluated the impact of the temperature coefficient $\tau$ of the WCE loss and the temperature coefficient $\tau_w$ of computing the weighted labels on the final performance of MSCOCO, and we depict these trends in Fig.~\ref{fig:para}(b) and (c). 

%-------------------------------------------------------------------------
\section{Conclusion}\label{sec:con}
In this paper, we propose a weighted contrastive hashing model to explore semantic information based on fine-grained information interactions between patches for image retrieval. The proposed mutual attention module can well solve the inconsistency of the anchor image and the augmented images. A weighted coefficient is calculated to weigh the similarities of the images in a training batch, and it can better improve the hash code learning. Extensive experiments show that the proposed method improves the state-of-the-art unsupervised hashing scheme in image retrieval.
%-------------------------------------------------------------------------

\subsubsection{Acknowledgements}
This work was supported in part by the National Natural Science Foundation of China (NSFC) under Grants No. 61872187, and No. 62072246, in part by the Natural Science Foundation of Jiangsu Province under Grant No. BK20201306, and in part by the ``111'' Program under Grant No. B13022.

 \bibliographystyle{splncs04}
 \bibliography{refs}

\end{document}